\documentclass[conference]{IEEEtran}
\IEEEoverridecommandlockouts
\usepackage{cite, array}
\usepackage{amsmath,amssymb,amsfonts}
\usepackage{algorithmic}
\usepackage{graphicx}
\usepackage{textcomp}
\usepackage{xcolor}
\def\BibTeX{{\rm B\kern-.05em{\sc i\kern-.025em b}\kern-.08em
    T\kern-.1667em\lower.7ex\hbox{E}\kern-.125emX}}
\begin{document}

\title{Forecasting Local Behavior of Self-organizing Many-agent System without Reconstruction\\
\thanks{This work was supported in part by the Office of Naval Research under Grant N00014-20-1-2432. The views and conclusions contained in this document are those of the authors and should not be interpreted as representing the official policies, either expressed or implied, of the Office of Naval Research or the U.S. Government.}

\author{\IEEEauthorblockN{Beomseok Kang, Minah Lee, Harshit Kumar, and Saibal Mukhopadhyay}
\IEEEauthorblockA{\textit{School of Electrical and Computer Engineering} \\
\textit{Georgia Institute of Technology}\\
Atlanta, USA \\
\{beomseok, minah.lee, hkumar64, smukhopadhyay6\}@gatech.edu}}}

\maketitle

\begin{abstract}
Large multi-agent systems are often driven by locally defined agent interactions, which is referred to as self-organization. Our primary objective is to determine when the propagation of such local interactions will reach a specific agent of interest. Although conventional approaches that reconstruct all agent states can be used, they may entail unnecessary computational costs. In this paper, we investigate a CNN-LSTM model to forecast the state of a particular agent in a large self-organizing multi-agent system without the reconstruction. The proposed model comprises a CNN encoder to represent the system in a low-dimensional vector, a LSTM module to learn agent dynamics in the vector space, and a MLP decoder to predict the future state of an agent. As an example, we consider a forest fire model where we aim to predict when a particular tree agent will start burning. We compare the proposed model with reconstruction-based approaches such as CNN-LSTM and ConvLSTM. The proposed model exhibits similar or slightly worse AUC but significantly reduces computational costs such as activation than ConvLSTM. Moreover, it achieves higher AUC with less computation than the recontruction-based CNN-LSTM.
\end{abstract}


\section{Introduction}
Artificial intelligence (AI) has been widely applied in multi-agent systems, both in synthetic environments and the real world \cite{vinyals2019grandmaster, kang2022unsupervised, chu2019multi, salzmann2020trajectron++}. One of the key challenges in implementing intelligent machines in such systems is to discover the hidden interaction rules between agents from sensory signals. Deep neural networks have emerged as powerful nonlinear tools for modeling the complex interaction rules governing these systems \cite{battaglia2016interaction}. However, learning the dynamics of large-scale multi-agent systems remains a challenging problem. Efficient deep learning models are needed to accelerate the progress of multi-agent applications, such as game AI, social networks, and robotics.

\begin{figure}
\begin{center}
\includegraphics[width=\columnwidth]{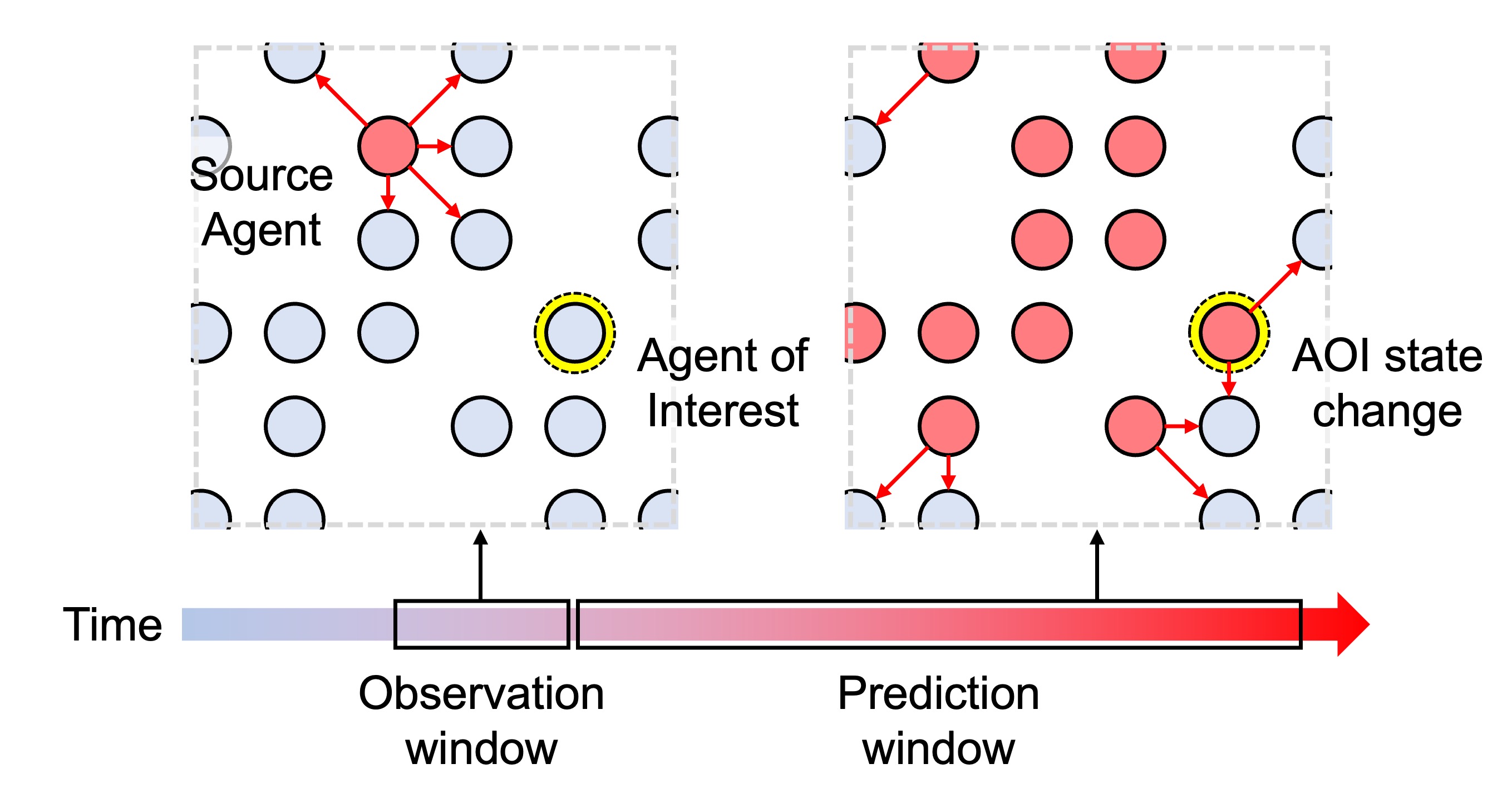}
\end{center}
\caption{Self-organizing many-agent system. Assume that each agent in the system is classified into binary states. Initial source agent affects neighboring agents based on pre-defined interaction rules and propagate through them. The prediction model aims to forecast how the state of an agent of interest (AOI) changes in a prediction window.}
\label{figure_intro}
\end{figure}

This paper focuses on many-agent systems with a large number of agents (thousands or more), where agents locally interact following non-linear dynamics. The global state of such system is defined by the collection of local interactions, which is often referred to as self-organization. Examples of such systems include the information diffusion in crowds, fire propagation in forests, and epidemic spreading between animals and humans \cite{albi2016invisible, zhou2018forestry, drossel1992self, shi2017voluntary}. Our focus, however, is on predicting the behavior of a few localized agents within these many-agent systems, assuming that other agents are beyond our concern as illustrated in Fig. 1. Previous works have developed AI models to learn interactions among a few agents (around 10 agents), assuming that all agents interact with each other \cite{saha2020magnet, battaglia2016interaction}. However, such approaches would be challenging particularly in self-organizing many-agent systems due to extremely high computational costs from agent-wise interactions and the lack of locality in the models, which makes training difficult. In summary, we aim to develop an efficient AI model for the novel task of predicting the behavior of a specific agent, which is defined as an agent of interest (AOI), in a self-organizing many-agent system.

\begin{figure*}[t]
\begin{center}
\includegraphics[width=\textwidth]{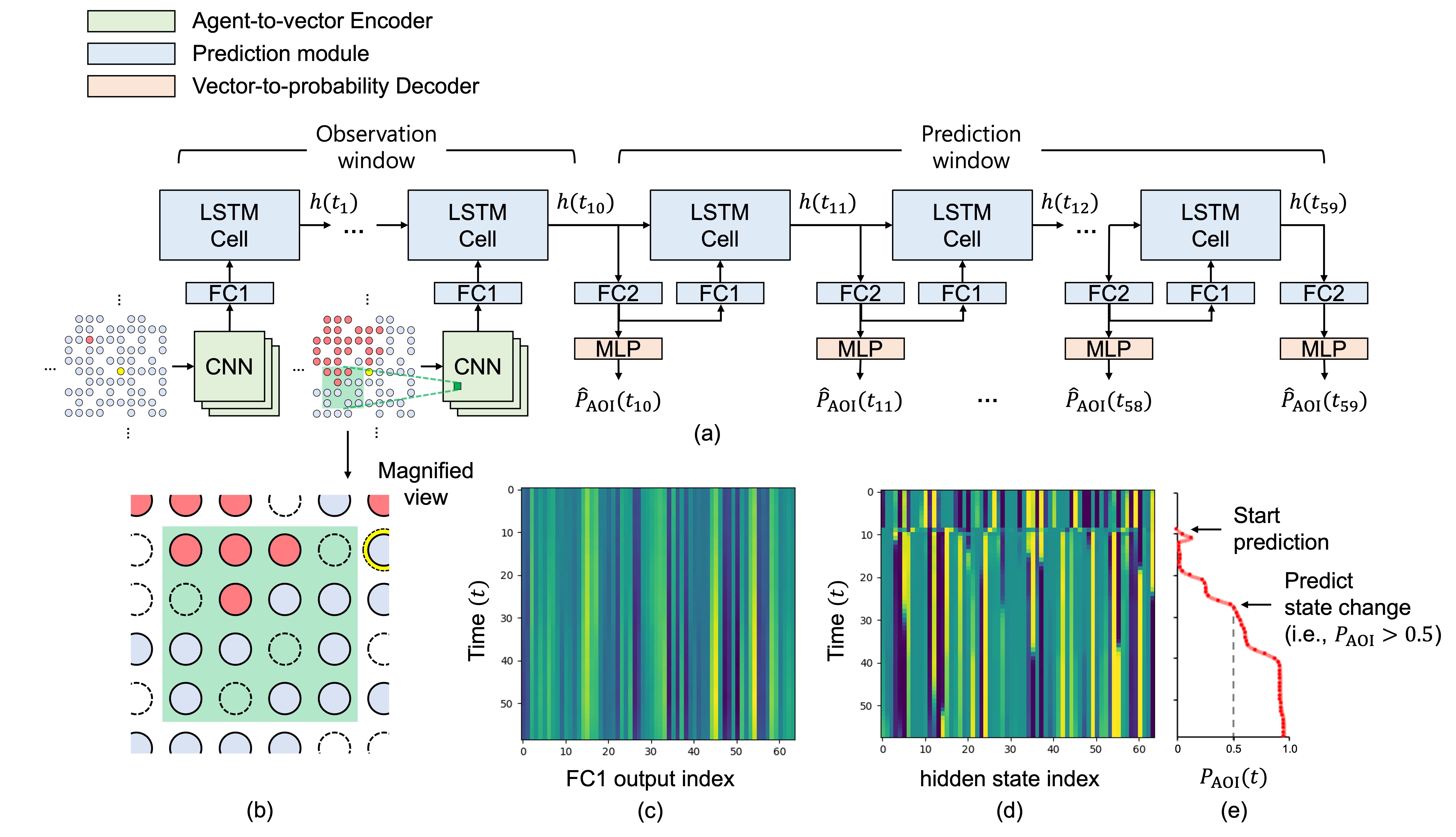}
\end{center}
\caption{AOI state prediction model. (a) data flow in the three modules: agent-to-vector encoder, prediction module, and vector-to-probability decoder. The encoder includes three convolution blocks with (7,7), (3,3), and (3,3) kernels with stride 2, batch normalization, ReLU activation, and max pooling layers with (3,3) kernels and stride 2. The LSTM has a single uni-directional cell with 64 hidden states, and the input and output of LSTM is processed by fully-connected layers. The decoder includes two full-connected layers with ReLU and Sigmoid activation. (b) magnified view of the encoder input. (c) encoder output for observed 60 timesteps. (d) hidden states for observed 10 timesteps and predicted 50 timesteps. (e) predicted probabilities for the 50 timesteps.}
\label{figure_model}
\end{figure*}

Self-organization in multi-agent systems is closely related to cellular automata. Recently, neural cellular automata has shown that complex self-organized systems can be modeled by convolutional neural networks (CNN) \cite{gilpin2019cellular, mordvintsev2020growing}. The proposed approaches are similar to autoencoders as they reconstruct the full image of the system to learn the transition of agent states. While the state of all agents including an AOI are easily accessible, the computational cost of the reconstruction is expensive in the large systems, and the temporal dynamics of agents would not be fully captured, which may limit their performance. Additionally, if the goal is to predict the local behavior of an agent, the reconstruction may not be necessary. This raises the question of whether AI surrogate models can predict the local behavior of many-agent systems without the need for reconstruction.

In this paper, we propose a CNN-LSTM model, which combines convolutional neural networks (CNN) and long short-term memory (LSTM), to predict the state of an AOI without the reconstruction \cite{hochreiter1997long}. Specifically, the proposed model predicts the state repeatedly at each time step in a prediction window after observing the state of all agents for a few sequential timesteps. The model is trained and evaluated on a forest fire model in NetLogo, which is a widely used agent-based modeling environment \cite{wilensky1999netlogo}. We observe that the proposed model makes an agent's state more predictable with less computation than a reconstruction-based model designed with the same encoder and prediction module. Also, we demonstrate that separately learning the spatial and temporal feature significantly saves computational costs such as the activation than convolutional LSTM (ConvLSTM) \cite{shi2015convolutional}.

\begin{figure*}[t]
\begin{center}
\includegraphics[width=\textwidth]{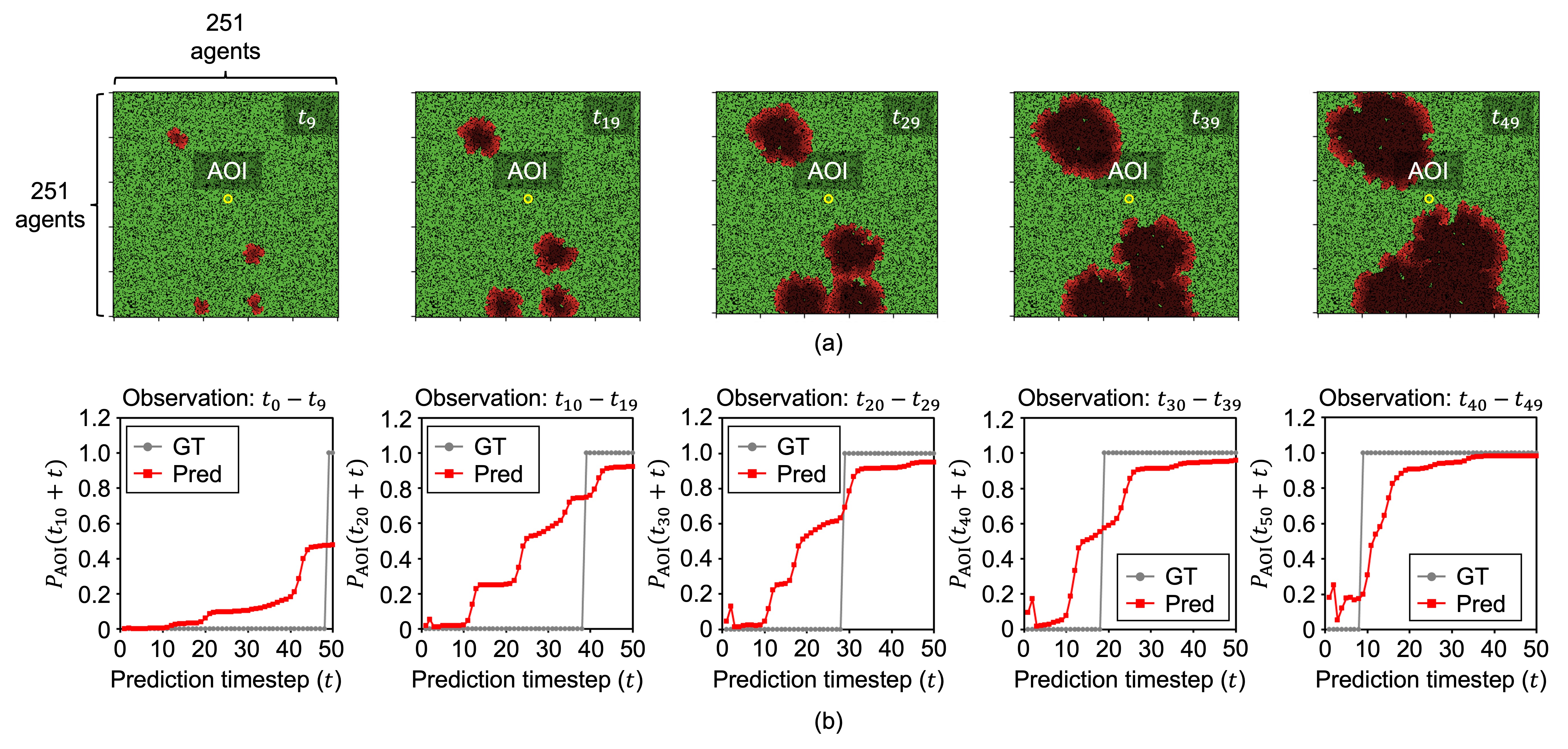}
\end{center}
\caption{Prediction for the burning probability of AOI. (a) forest state at the last timestep of a observation window. (b) predicted burning probabilities (Pred) and ground truth (GT) in a corresponding prediction window. The observation and prediction window shift by 10 timesteps from left to right graphs. For example, the model in the left first graph observes the first 10 timesteps (\(t_{0} - t_{9}\)) and predicts the next 50 timesteps (\(t_{10} - t_{59}\)). Forest density in the figure and graph is 76.}
\label{figure_prob}
\end{figure*}

\section{Proposed Approach}
\subsection{Prediction Model Architecture}

Our prediction model consists of three principal modules: agent-to-vector encoder, prediction module, and vector-to-probability decoder. The architecture is described in Fig. \ref{figure_model}(a). The CNN-based encoder transforms a set of agents into a context vector during a observation window, and then the LSTM-based prediction module accumulates the temporal information of agents into hidden states, \(h(t)\). FC1 and FC2 in the figure refer fully-connected layers to match the dimension of the CNN output and LSTM input. The LSTM predicts the change in a AOI's state as a probability, \(P_{\text{AOI}}\), every time step in a prediction window through the MLP-based decoder. The recursive prediction of the LSTM in a prediction window is based on the previously predicted latent vectors. The detailed configuration of each layer is noted in the figure caption.

We assume that agents are randomly distributed in 251 by 251 areas, where each agent takes 1 by 1, and agents are spatially stationary which makes convolution operations available to the groups of neighbor agents. Fig. \ref{figure_model}(b) illustrates neighbor agents processed by the CNN encoder. As agents can be spatially sparse, empty areas are considered empty agents, which are represented by zero feature vectors. In a forest fire model, we represent each of non-empty agents, such as trees and fires, into 1 by 3 color vectors (i.e., RGB) while empty agents are represented by zeros (i.e., black in terms of RGB). 

To illustrate the inner working of the model, Fig. \ref{figure_model}(c), (d), and (e) show the output of the CNN encoder, hidden states in the LSTM, and the predicted \(P_{\text{AOI}}\), respectively. Note that the CNN output in Fig. \ref{figure_model}(c) is the encoded results for 60 timesteps in the environment for the illustration purpose, and the prediction model only uses the encoder for the first 10 timesteps. That is, the first 10 timesteps are observed results and the rest 50 timesteps are predicted ones in Fig. \ref{figure_model}(d) and (e). The figures are based on a forest fire model, which we will discuss in detail in the following sections. We observe that the encoder output is slowly changing as fire propagates through agents, whereas the hidden states and probabilities are relatively dynamic in a prediction window. It implies that the encoder aggregates the global feature of agents given the insensitive changes to an AOI state. In contrast, the dynamics of the hidden states highly depend on an AOI state, which will facilitate decoding the probability of a specific local agent from a large number of agents.

A key advantage of the proposed model is low computational costs. Given other spatiotemporal learning models such as ConvLSTM internally process hidden states as voxels, their computational costs are often expensive due to large activation from hidden states. On the other hand, our model has the separated spatial and temporal learning modules and creates 1D hidden states, thereby significantly saving the computational cost. Moreover, the LSTM can predict the context vector repeatedly using previously predicted ones, which enables the direct prediction of a particular agent's state from the vector without the need for reconstruction.

\subsection{Training Procedure}
We define the first 10 and next 50 timesteps as an observation and prediction window, respectively. The prediction module accumulates temporal information from the context vectors in an observation window and predicts the next context vectors through an autoregressive process. In other words, the model observes the first 10 timesteps and generates the probabilities, \(\hat{P}_{\text{AOI}}\) for each of the next 50 timesteps. The model is trained to minimize the binary cross-entropy (BCE) loss:
\begin{equation}
\begin{split}
L({P_{\text{AOI}}, \hat{P}_{\text{AOI}}})= -\frac{1}{N}\sum_{t=0}^{N-1} & {P_{\text{AOI}}(t)} \log(\hat{P}_{\text{AOI}}(t)) + \\
& (1-{P_{\text{AOI}}(t)}) \log(1-\hat{P}_{\text{AOI}}(t))
\end{split}
\end{equation}
\noindent where \(N\) is the length of a prediction window. Hence, the average loss for the 50 probabilities is optimized through backpropagation through time. We use an Adam optimizer and set the learning rate to 5e-6, train batch size to 4, and the number of epochs to 100. We refer \(P_{\text{AOI}}\) as a burning probability from here.

\section{Experimental Result}
\subsection{Dataset: Agent-based Forest Fire Model}
We utilize an agent-based model to train and evaluate the proposed prediction model, using a forest fire as an example of a self-organizing many-agent system. The forest fire model consists of three types of agents: fire, ember, and tree, and their interactions result in emergent behavior of fire propagation. The agents are classified into two binary states: 0 for not burning (i.e., tree) and 1 for burning (i.e., fire and ember). The prediction model is trained to output a high burning probability if an AOI is in a burning state.

The forest fire model involves three key phases: (1) fire seeds start at random locations, (2) fire evolves from the seeds, and (3) fire stops spreading. The model is driven by a set of interaction rules, which is inspired by the Rothermel equation that describes the evolution of the forest fire \cite{rothermel1972mathematical}. The initial condition of the simulation is defined as follows:
\begin{equation}
Q_{(i,j)}=
    \begin{cases}
      I_{\text{seed}} \times Q_{\text{th}} & \text{if} \: (i,j) \in \text{seeds} \\
      0 & \text{otherwise}
    \end{cases}
\end{equation}


\noindent where $Q_{(i,j)}$ represents the heat accumulation at an agent's location $(i,j)$, \(I_{\text{seed}}\) is the intensity of initial fire seeds, and $Q_{\text{th}}$ is a threshold value that determines the condition for ignition. The heat accumulation for unburned agents is determined by:
\begin{equation}
Q_{(i,j)}(t+\Delta t) = Q_{(i,j)}(t) + \lambda \sum_{(k,l) \in N_R} \mathbf{1}_{(k,l)}(t) Q_{(k,l)}(t)
\end{equation}

\noindent where \(\lambda\) is transfer efficiency, which determines the efficiency of heat transfer from a set of neighboring agents, $N_R$, and $\mathbf{1}_{(k,l)}(t)$ is an indicator function that ensures agents on fire can only contribute to the heat accumulation. A fire agent will ignite and convert to a burning state if $Q_{(i,j)}(t) > Q_{\text{th}}$. A fire agent converts to an ember agent in the next time step and undergoes heat decay which is given by:
\begin{equation}
Q_{(i,j)}(t+\Delta t) = Q_{(i,j)}(t) - Q_{\text{die}}
\end{equation}

\noindent where $Q_{\text{die}}$ is the rate at which the embers dissipate heat. There are multi parameters that affect fire propagation, and implementation details, including these parameters, can be found online\footnote{https://github.com/harshitk11/NetLogo-Forest-Fire-evolution}. One of the critical parameters that directly impacts fire evolution is forest density. For instance, a high-density forest leads to a radially diffusing forest fire, whereas fire paths are more chaotic in a low-density forest.

To train and evaluate the proposed model, we generated chunk-based training and test datasets with multiple simulations. The tree distribution and location of initial fire seeds are randomly selected for each simulation. Each chunk comprises successive 60 timesteps of the forest, and multiple chunks for each simulation are generated with a 10-timestep difference. For instance, the first and second chunks define the 0-th to 59-th timesteps and 10-th to 69-th timesteps, respectively. We considered two density parameters, 76 and 72, to study the model's performance in different forest settings. The density number indicates that the percentage number of agents in the forest. A higher density parameter indicates that there are more trees in the forest. For our density 76 datasets, we included 970 chunks (70 simulations) for training and 1386 chunks (100 simulations) for testing. For the density 72 datasets, we included 1255 chunks (70 simulations) for training and 912 chunks (50 simulations) for testing.

\begin{figure}[t]
\begin{center}
\includegraphics[width=\columnwidth]{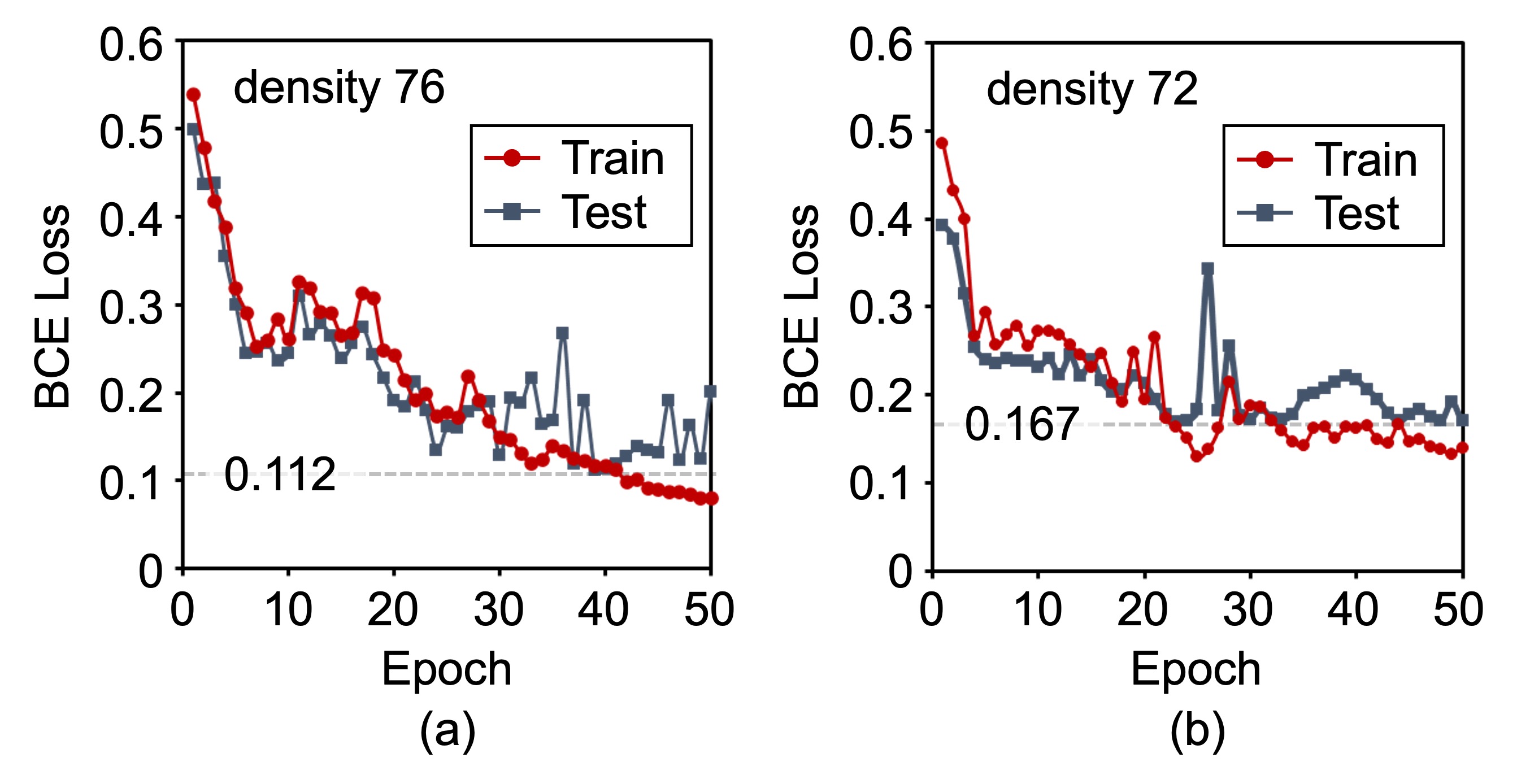}
\end{center}
\caption{(a) Train and test loss in density 76 and (b) density 72. The model arrives at the minimum test loss within 50 training epochs. The minimum test loss is noted in the figure.}
\label{figure_loss}
\end{figure}

\begin{figure}[t]
\begin{center}
\includegraphics[width=\columnwidth]{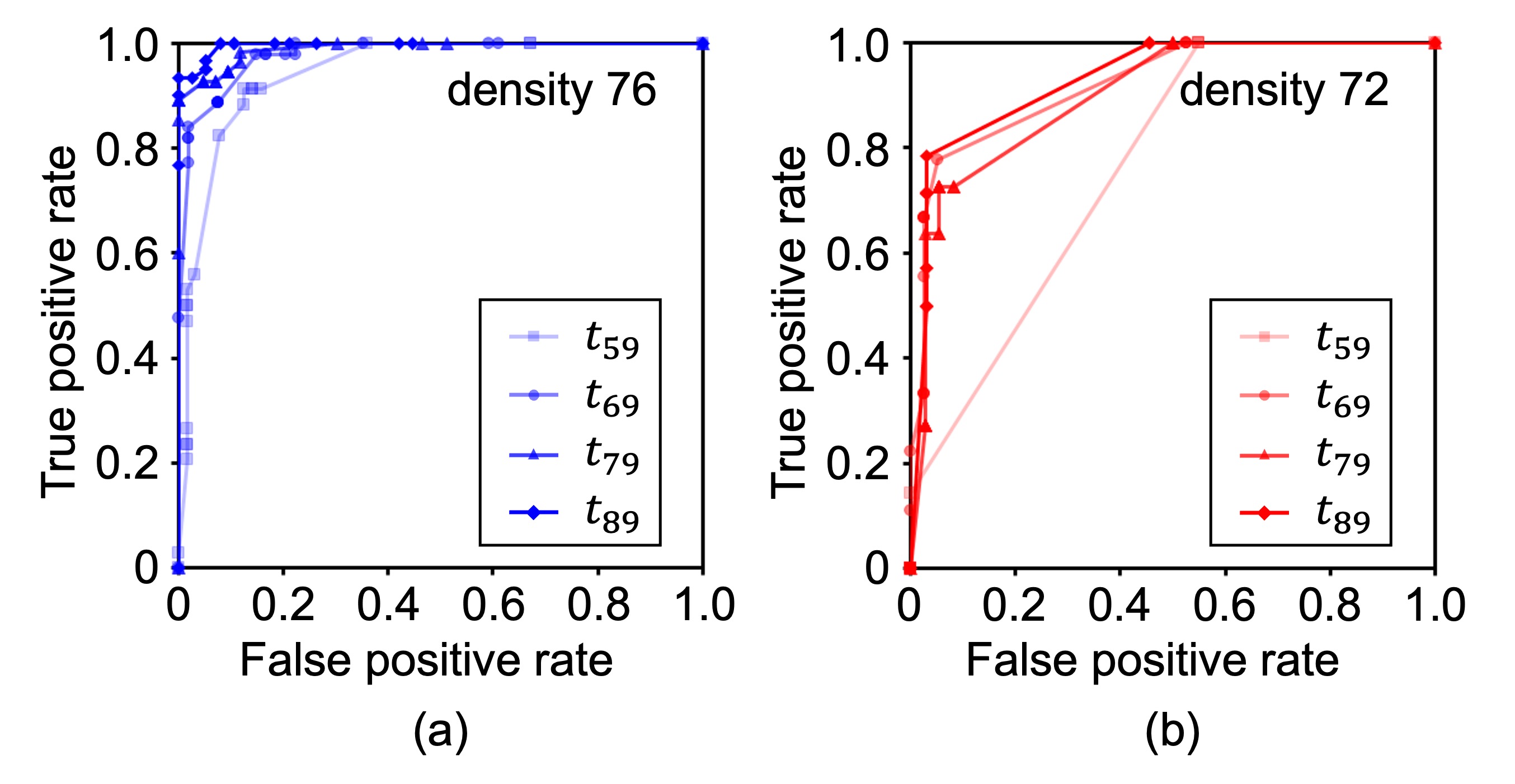}
\end{center}
\caption{(a) ROC curve in density 76 and (b) density 72. The total 21 uniform decision thresholds, such as 0.0, 0.05, 0.1, are considered to define positive and negative case from the model output.}
\label{figure_roc}
\end{figure}

\subsection{Burning Probability Prediction for AOI}
The two models are trained individually on densities 72 and 76 to predict the burning probability of an AOI at the coordinate (125, 125), which is the center of a forest. The coordinate follows that the left top corner in the forest is (0, 0) and the right bottom corner is (250, 250). Fig. \ref{figure_prob}(b) shows the predicted and ground truth burning probabilities (\(p_{\text{burn}}\)) of the AOI while shifting the observation and prediction window by 10 time steps. The predicted probabilities gradually increase when the fire event of the AOI occurs later in the prediction window, while they sharply change if the fire occurs in the near future. Although the model has shown reasonable prediction results for the density 76, we observe noticeable performance degradation in the low-density forest. Fig. \ref{figure_loss} displays the training and test loss for 50 epochs. The minimum test loss is noted on both the graphs, indicating the low-density forest is less predictable. 

Fig. \ref{figure_roc} presents the receiver operating characteristics (ROC) curves at the first four chunks in the two densities. A positive case indicates that the AOI is burning and \(\hat{P}_{\text{AOI}}\) is higher than 0.5 for the last timestep in the prediction window, such as the 59-th timestep in the first chunk. We observe that the model is sensitive to probability thresholds in the low-density forest, particularly failing at the first chunk. In other words, the fire evolution is more chaotic in the low-density forest. Given that fire propagation paths are determined by the existence of trees, fires in the high-density forest are likely to radially diffuse, whereas fires in the low-density forest involve more complex paths, which do not necessarily follow a radial direction from a fire seed. Also, the ground truths are binary, and the model's prediction can be considered the repeated binary classification of an AOI's state. The classification becomes easier if the forest with a burning AOI share common features, such as large burning areas near the AOI. However, the low-density forest is likely to entail less common features due to such complex propagation paths, thereby hindering the model from designing a classification boundary for the local behavior using the global feature of a forest.

\begin{figure}[t]
\begin{center}
\includegraphics[width=\columnwidth]{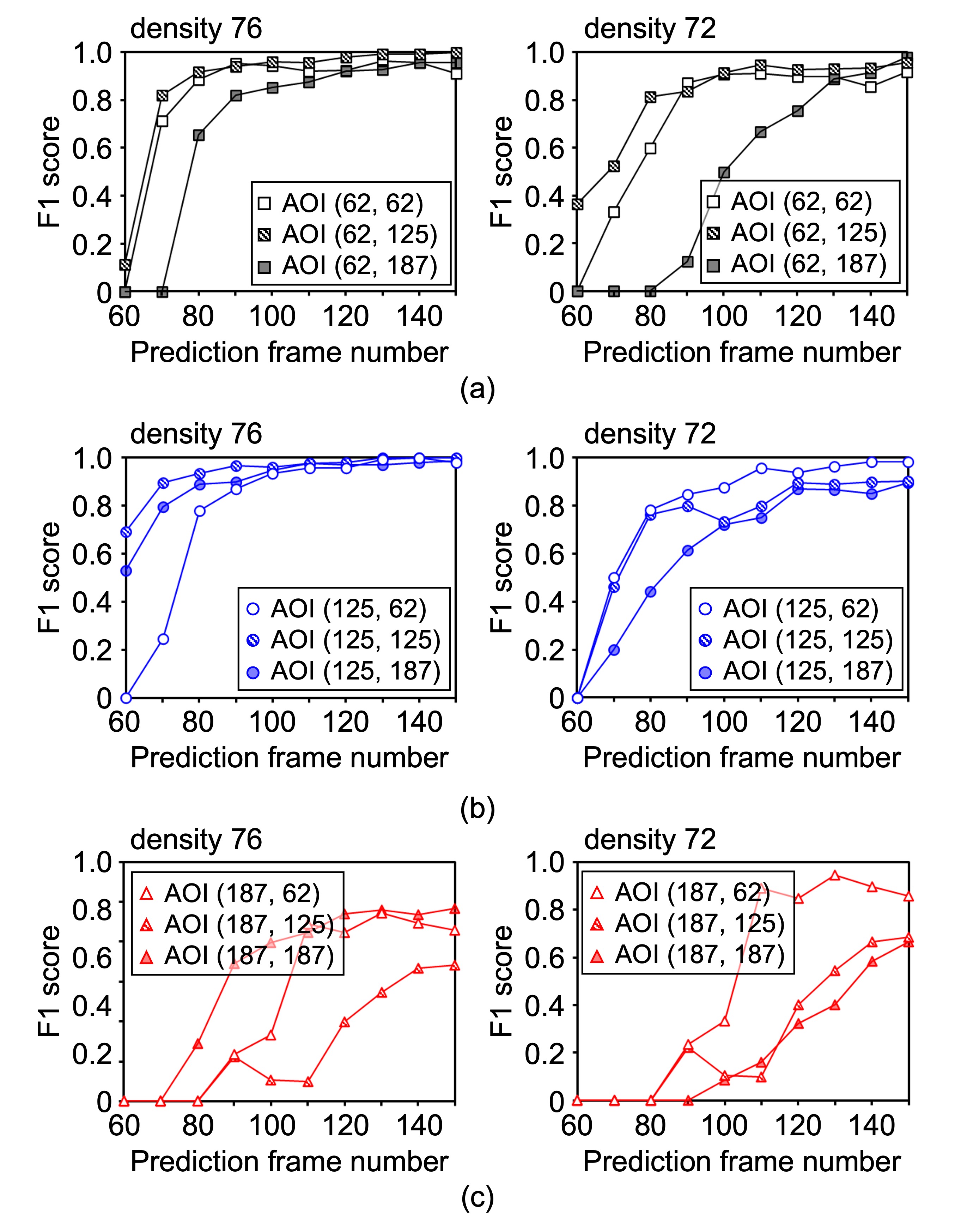}
\end{center}
\caption{F1 score for multi AOIs. (a) F1 score for AOIs of north area, (b) middle area, (c) and south area in both the density 76 and 72. The decision threshold considered in F1 score is 0.5.}
\label{figure_roi}
\end{figure}

\subsection{Burning Probability Prediction for multi AOIs}
The burning probability prediction is performed for multi AOIs with the models individually trained for each AOI. 9 AOIs are chosen to be uniformly distributed in the forest. Specifically, each of the AOIs is the center agent when we divide the forest into 9 patches. The decision threshold for the predicted burning probability is also set to 0.5, and the last timestep in the prediction window, such as the 60-th timestep for the first chunk, is considered for the evaluation. Fig. \ref{figure_roi} shows F1 score for the AOIs. Again, the coordinate of the left top corner in the forest is (0, 0) and the right bottom corner is (250, 250). We observe that the performance of the proposed models heavily depend on the AOIs in both the densities. Also, most of the cases exhibit that the F1 score is particularly low at the first chunk (i.e., the 60-th timestep) and gradually increase in later timesteps.

\begin{figure}[t]
\begin{center}
\includegraphics[width=\columnwidth]{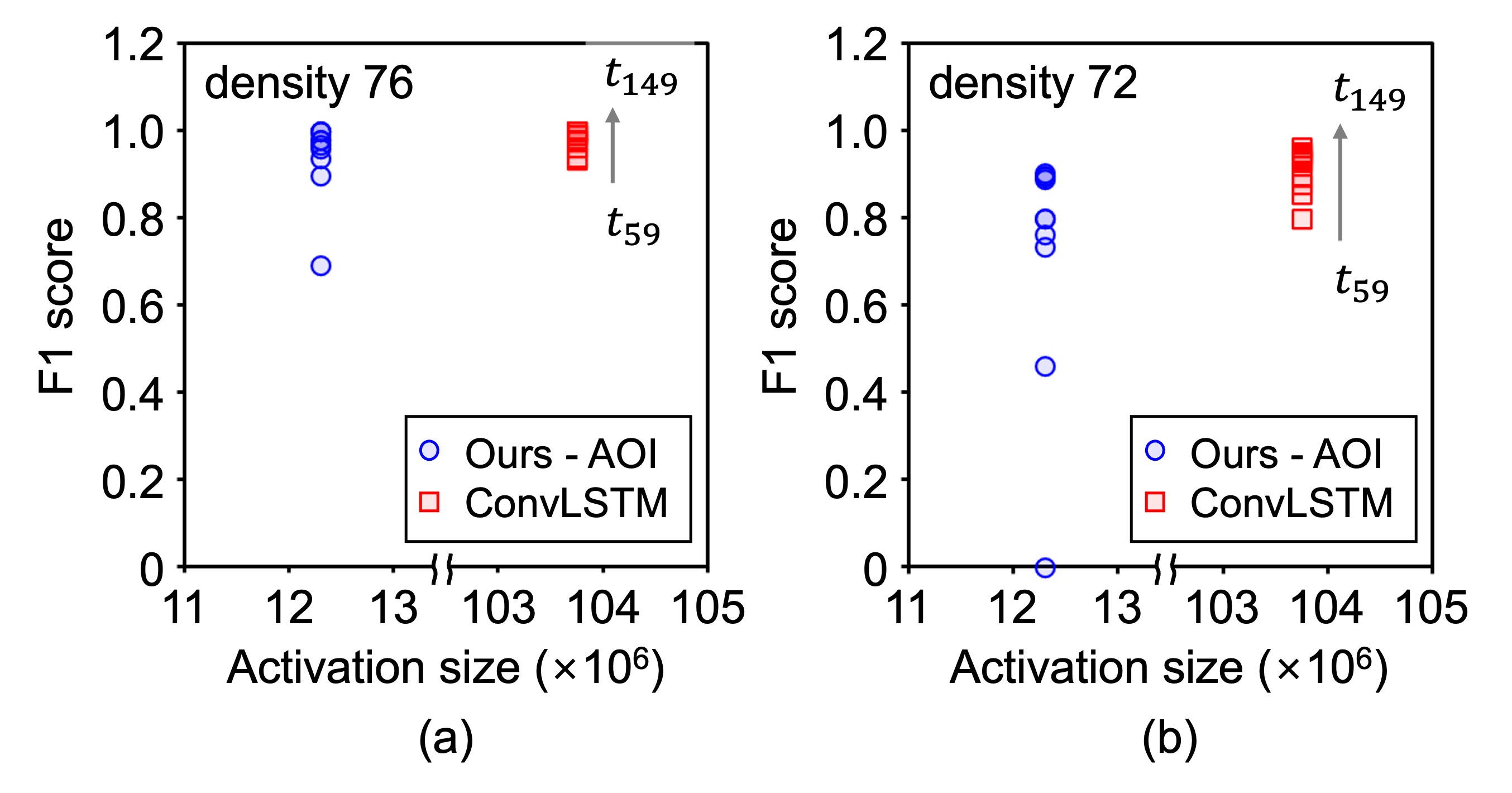}
\end{center}
\caption{F1 score with activation size. (a) Comparison of F1 score and activation size between the proposed model and ConvLSTM in density 76 and (b) density 72. The data points at lower F1 score indicates earlier prediction windows. The decision threshold considered is 0.5.}
\label{figure_convlstm}
\end{figure}

\subsection{Comparison with Reconstruction-based Model}
We compare the proposed model to ConvLSTM with the similar amount of trainable parameters. The ConvLSTM model generates an probability map of the forest, instead of a particular AOI. We implement an encoder with the first convolutional block in the proposed model, a single-layer ConvLSTM cell with (3,3) kernel and (64, 61, 61) hidden and cell states, and a decoder with three upconvolutional blocks with (3,3) kernels. Fig. \ref{figure_convlstm} shows the F1 score for an AOI at (125, 125) versus the total activation during the prediction. The scatters in the figure indicate the F1 score for the last timestep in the different prediction windows. The ConvLSTM model shows the descent performance throughout the overall chunks while the proposed models fail to achieve the high F1 score in the early prediction windows. However, it is important to note the activation of ConvLSTM is much larger than the proposed model. This is mainly because the large hidden state in the ConvLSTM cell dominates the activation while the hidden state of the proposed model is a small 1D vector.

\begin{table}
\centering
\caption {Computational cost of the models.}
\begin{tabular}{>{\centering\arraybackslash}p{2cm}>{\centering\arraybackslash}p{1.5cm}>{\centering\arraybackslash}p{1.5cm}}
\hline
\\ [-0.9em]
Model & Parameter & Activation \\
\\ [-1.1em]
\hline
\\ [-0.9em]
Ours - AOI & 262.7k & 12.3M \\
\\ [-1.1em]
Ours - Reconst. & 295.6k & 12.8M \\
\\ [-1.1em]
ConvLSTM & 250.6k & 103.8M \\
\\ [-1.1em]
\hline
\end{tabular}
\label{table_comp}
\end{table}

\begin{table}
\centering
\caption {AUC of the models for AOI (125, 125) in density 76 (top) and density 72 (bottom).}
\begin{tabular}{>{\centering\arraybackslash}p{2cm}>{\centering\arraybackslash}p{1cm}>{\centering\arraybackslash}p{1cm}>{\centering\arraybackslash}p{1cm}>{\centering\arraybackslash}p{1cm}}
\hline
\\ [-0.9em]
Model & AUC (\(t_{59}\)) & AUC (\(t_{69}\)) & AUC (\(t_{79}\)) & AUC (\(t_{89}\)) \\
\\ [-1.1em]
\hline
\\ [-0.9em]
Ours - AOI & 0.946 & 0.979 & 0.990 & 0.996 \\
\\ [-1.1em]
Ours - Reconst. & 0.730 & 0.906 & 0.952 & 0.985 \\
\\ [-1.1em]
ConvLSTM & 0.970 & 0.989 & 0.998 & 1.000 \\
\\ [-1.1em]
\hline\hline
\\ [-0.9em]
Model & AUC (\(t_{59}\)) & AUC (\(t_{69}\)) & AUC (\(t_{79}\)) & AUC (\(t_{89}\)) \\
\\ [-1.1em]
\hline
\\ [-0.9em]
Ours - AOI & 0.764 & 0.921 & 0.902 & 0.932 \\
\\ [-1.1em]
Ours - Reconst. & 0.748 & 0.757 & 0.942 & 0.936 \\
\\ [-1.1em]
ConvLSTM & 0.970 & 0.971 & 0.972 & 0.970 \\
\\ [-1.1em]
\hline
\end{tabular}
\label{table_auc}
\end{table}

Table \ref{table_comp} and \ref{table_auc} provide a summary of the computational cost and area under the ROC curve (AUC) of the models we evaluated. The "activation" column in Table \ref{table_comp} shows the total activation dimension required for predicting 50 time steps. We also design a reconstruction-based model (i.e., Ours-Reconst. in the table), which uses the same encoder and prediction module as the CNN-LSTM model, but with a modified decoder. The reconstruction-based model's decoder consists of upconvolutional blocks, each of which includes a (2,2) upsample, a convolutional layer, batch normalization, and ReLU activation. All convolutional layers have a (3,3) kernel with stride 1 and padding 1. While the CNN-LSTM and ConvLSTM models have similar numbers of parameters, the CNN-LSTM models have much lower activation due to the use of a small vector to learn the dynamics. The AOI-based model shows the lowest activation because its decoder does not reconstruct the states of all agents. Furthermore, it achieves higher AUC scores than the reconstruction-based model in most prediction windows. Although the ConvLSTM model achieves the highest AUC scores, the AOI-based model has comparable performance in density 76.

One possible explanation for the better performance of the ConvLSTM is that it conserves the forest's spatial structure better. The ConvLSTM's hidden states have large width and height dimensions than those of the proposed CNN-LSTM model. As a result, hidden state updates in the ConvLSTM are associated with more localized areas in the forest, whereas in the CNN-LSTM, such updates affect larger areas, resulting in a loss of local information. The poor performance of the reconstruction-based CNN-LSTM model supports this hypothesis, as we observe that it fails to accurately recover the forest from its hidden states. However, since self-organizing multi-agent systems involves locally defined interaction, the preservation of local information in the ConvLSTM is an important consideration, despite its approximately 9 times larger (or possibly even larger for longer predictions) activation compared to the CNN-LSTM. Ultimately, there is a trade-off between the conservation of the system's spatial structure and the computational cost resulted from the large dimension of the hidden states.

\section{Conclusion}
In this paper, we have proposed a CNN-LSTM model for predicting the state of an agents without the reconstruction in a self-organizing many-agent system. Our model was evaluated in a NetLogo environment, and the results showed that the AOI-based model outperforms the reconstruction-based CNN-LSTM model for certain AOIs, achieving higher F1 score and AUC with less computation. We have also compared our evaluation metric with the ConvLSTM model, and while the proposed model exhibits a lower AUC, it significantly saves computational cost such as activation. Our work highlights the potential of a novel architecture that can preserve the spatial structure of a system and learn its dynamics without significantly increasing computational costs, particularly for large self-organizing multi-agent systems.

\bibliographystyle{IEEEtran}

\bibliography{refs.bib}

\end{document}